# DYNAMIC REGION PROPOSAL NETWORKS FOR SEMANTIC SEGMENTATION IN AUTOMATED GLAUCOMA SCREENING

*Shivam Shah, Nikhil Kasukurthi, Harshit Pande*

SigTuple Technologies Pvt. Ltd., Bengaluru, India

## ABSTRACT

Screening for the diagnosis of glaucoma through a fundus image can be determined by the optic cup to disc diameter ratio (CDR), which requires the segmentation of the cup and disc regions. In this paper, we propose two novel approaches, namely Parameter-Shared Branched Network (PSBN) and Weak Region of Interest Model-based segmentation (WRoIM) to identify disc and cup boundaries. Unlike the previous approaches, the proposed methods are trained end-to-end through a single neural network architecture and use dynamic cropping instead of manual or traditional computer vision-based cropping. We are able to achieve similar performance as that of state-of-the-art approaches with less number of network parameters. Our experiments include comparison with different best known methods on publicly available Drishti-GS1 and RIM-ONE v3 datasets. With $7.8 \times 10^6$ parameters our approach achieves a Dice score of $0.96/0.89$ for disc/cup segmentation on Drishti-GS1 data whereas the existing state-of-the-art approach uses $19.8 \times 10^6$ parameters to achieve a dice score of $0.97/0.89$.

***Index Terms*—** deep learning, image segmentation, optic disc & cup, glaucoma screening

## 1. INTRODUCTION

Glaucoma is an ocular disease that causes damage to the optic nerve with ensuing vision loss [1]. It is the second leading cause of progressive irreversible vision loss in the world [2]. Early detection and monitoring with timely treatment can mitigate serious vision loss [3]. As shown in Figure 1, ophthalmologists mark both the optic disc margin at the inner edge of the scleral ring and the white optic cup in the centre of the disc. The ratio of the optic cup diameter to that of the disc diameter is a useful metric for an indication of glaucoma. Automated screening of early stage glaucoma by the segmentation of the optic disc and the cup in a fundus photograph is an active area of research. In order to assist ophthalmologists, various computer vision and deep learning-based approaches have been proposed for segmenting the optic disc and the cup region.

Earlier, segmentation of the optic disc and the cup was mainly done through the traditional computer vision tech-niques. A comparative study on various such techniques is done by Almazroa *et al.* [4], which concludes that these methods are not robust to the changes in fundus camera or other image characteristics. Recently, deep learning-based approaches, which are more robust, have been applied for such segmentation tasks. Ronneberger *et al.* [5] introduce U-Net, a combination of encoding and decoding convolution neural network architectures, which is commonly used for biomedical image segmentation. For glaucoma-specific segmentation, Fu *et al.* [7] propose a modified U-Net architecture with multi-scale input/output at each convolution/deconvolutional block of the U-Net. The input is an image cropped to the disc region identified by a computer vision technique mentioned in [8]. Sevastopolsky [6] report performance of two separate U-Net models trained for disc and cup respectively. Training of the cup model is done with an assumption that the disc segmentation is known and hence the input image for the cup model is a cropped fundus image based on the disc region. Chakravarty *et al.* [9] use a single U-Net network with the classification of both disc and cup through separate loss functions. Here too the input image is the disc region estimated using a traditional computer vision technique described in Chakravarty *et al.* [10]. Sevastopolsky *et al.* [11] propose a stacked U-Net architecture in which they use a U-Net network as a basic block and stack 15 such basic blocks to achieve a higher recognition quality. In their approach, two separate stacked U-Net models are trained for disc and cup segmentation causing a linear increase in parameters as well as the inference time.

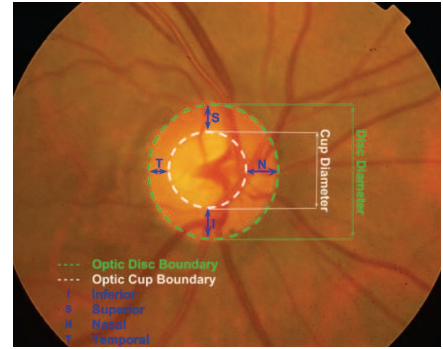

**Fig. 1**: Fundus photograph for glaucoma screening: ophthalmologist manually calculate the vertical cup to disc diameter ratio as well as ISNT metric which is a very tedious and time-consuming process [6].

The **key contribution** of this paper is a novel end-to-end deep learning architecture for the segmentation of the optic disc and the cup from a fundus photograph. In an integrated architecture, we first weakly predict the disc to estimate the region of interest (RoI) and then dynamically crop that region to train a U-Net with a three-channel output for background, cup, and disc rim. We evaluate our proposed methods on the publicly available Drishti-GS1 [12] and RIM-ONE v3 [13] datasets. On these datasets, we also compare our methods with the existing deep learning-based segmentation methods. The proposed approaches achieve results similar to those of the state-of-the-art solutions and yet have an advantage in terms of reducing the network parameters.

## 2. PROPOSED METHODS

### 2.1. Parameter-Shared Branched Network (PSBN)

In this section, we present a deep neural network architecture using shared parameters within two branches to learn the optic disc and the cup masks as shown in Figure 2. The first branch is a U-Net [5] like model which takes an RGB fundus photograph of a fixed $512 \times 512$ size as input. It is passed through an encoding and decoding network that predicts the optic disc mask. The second branch is for the optic cup mask, with shared encoding architecture and parameters with the first branch but a separate decoding architecture. In the second branch, we use the cropped activations from the encoder of the first branch. The cropping is done on the basis of the prediction of the disc mask from the first branch. The dynamic localization of disc prediction area is estimated using the centroid calculation from the image moment. If the disc mask prediction is represented by matrix $M_{disc}$ ($M_{disc}[i,j] \in 0, 1$), the location for cup mask can be approximated as a cropping of size $256 \times 256$ centered at:

$$R_{M_{cup}} = \frac{\sum_{i=0}^{N-1} \left\{ \sum_{j=0}^{N-1} M_{disc}[i,j] \right\} * i}{\sum_{i=0}^{N-1} \sum_{j=0}^{N-1} M_{disc}[i,j]} \quad (1)$$

$$C_{M_{cup}} = \frac{\sum_{j=0}^{N-1} \left\{ \sum_{i=0}^{N-1} M_{disc}[i,j] \right\} * j}{\sum_{i=0}^{N-1} \sum_{j=0}^{N-1} M_{disc}[i,j]} \quad (2)$$

where $N = 512$ and $R_{M_{cup}}, C_{M_{cup}}$ are the row and column values of the cup cropped centre in $M_{disc}$ matrix respectively. Now this cropped region corresponds to the disc prediction mask of size $512 \times 512$. Hence cropping of activation map $A_n$ (the cropped activations from the encoder) of size $n \times n$ can be evaluated as centered at:

$$R_{A_n} = \frac{R_{M_{cup}} \times n}{N}, C_{A_n} = \frac{C_{M_{cup}} \times n}{N} \quad (3)$$

where $R_{A_n}, C_{A_n}$ are the rows and column values of the cropped centre in the activation $A_n$. The size of this cropping is $\frac{n \times 256}{N}$. These cropped activations are concatenated to the corresponding upsampling layers of the cup segmentation branch as shown in Figure 2. The final output of second branch is padded with zeros to get the final cup segmentation.

### 2.2. Weak Region of Interest Model-based Segmentation (WRoIM)

This proposed network is inspired from the network described in the section 2.1, in which we observe that the performance gain for cup segmentation is significantly better than that for disc segmentation. Given the fact that in section 2.1, we use the disc prediction output to dynamically crop the activations for cup segmentation, which lead to improved performance of cup segmentation, we propose a network through which both the disc and the cup segmentation will be done on dynamically cropped fundus photograph. As shown in Figure 3, first we have a small U-Net architecture (1 conv block downsample and 1 conv block upsample) to weakly predict the disc region. Based on the weak predictions of the disc region, we dynamically crop the fundus image and pass it through the larger U-Net architecture to get a multi-channel prediction for background, cup, and disc rim (the area between disc and cup boundary). The rationale for adding a small U-Net is to first get an approximate region of interest (RoI) and then crop $256 \times 256$ sized box centered at RoI from fundus image as described in Equation 1 and 2.

## 3. EXPERIMENTAL SETUP

All the experiments are conducted on the publicly available Drishti-GS1 [12] and RIM-ONE v3 [13] datasets, for which the data distribution is shown in Table 1. As a pre-processing step, similar to Sevastopolsky *et al.* [6], we are also using Contrast Limited Adaptive Histogram Equalization (CLAHE) technique which enhances the contrast of an image. All the proposed approaches are implemented in Keras (with Tensorflow backend) and trained and tested on an Ubuntu machine having 16 core Intel(R) Xeon(R) CPU E5-2673 v3 @ 2.40GHz, 55 GB Memory and Nvidia Tesla M60 GPU.

| Source | Train | Test |
|---|---|---|
| Drishti-GS1 | 50 | 51 |
| RIM-ONE v3 | 100 | 59 |

**Table 1**: Data Distribution

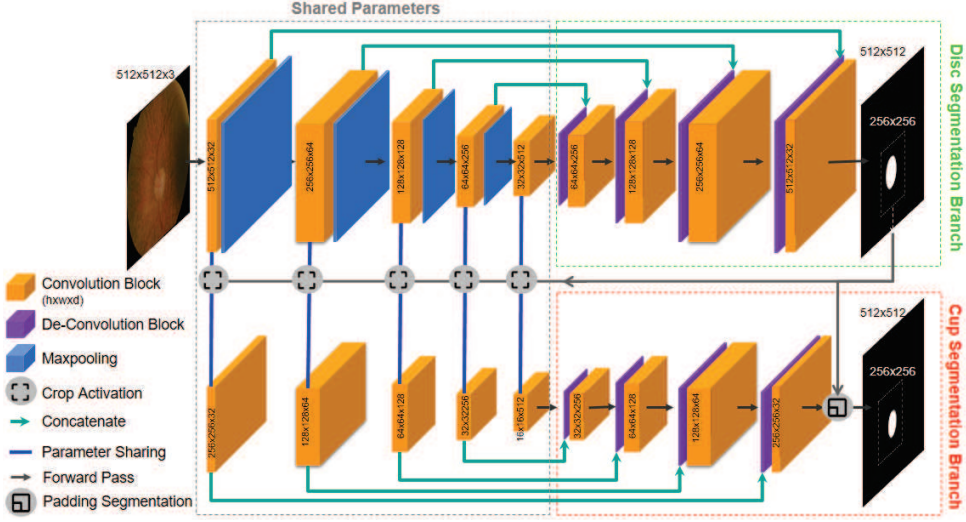

**Fig. 2**: Parameter-shared branched network

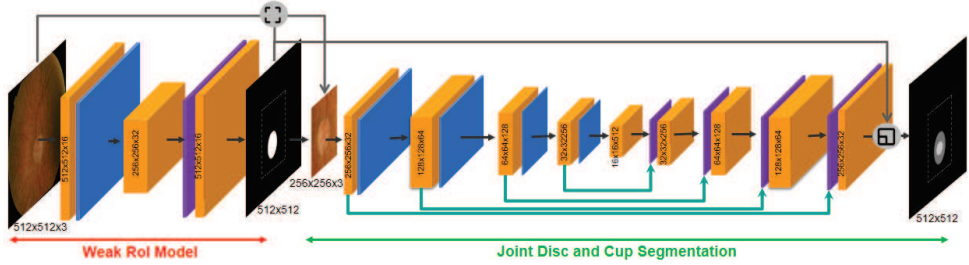

**Fig. 3**: Weak region of interest model-based segmentation network (all notations are the same as mentioned in Figure 2).

For a comparative study to understand the impact of region cropping, we also trained a 2 model network, which consists of two individual U-Nets. Disc and cup are trained independently with their respective ground truth masks. The input in both the cases is a full fundus image without any cropping of RoI.

For all three networks we are using logarithmic dice loss defined as:

$$Loss(X, Y) = -\log Dice(X, Y) \qquad (4)$$

$$Dice(X,Y) = \frac{2 \times \sum_{i=0}^{N-1} \sum_{j=0}^{N-1} \{x_{ij} \times y_{ij}\}}{\epsilon + \sum_{i=0}^{N-1} \sum_{j=0}^{N-1} \{x_{ij}^2\} + \sum_{i=0}^{N-1} \sum_{j=0}^{N-1} \{y_{ij}^2\}} \qquad (5)$$

where $X = [x_{ij}]_{i=0}^{N-1}{}_{j=0}^{N-1}$ with $x_{ij} \in {0, 1}$ represents the binary segmentation of the ground truth and $Y = [y_{ij}]_{i=0}^{N-1}{}_{j=0}^{N-1}$ with $y_{ij} \in [0, 1]$ is the prediction. $\epsilon = 10^{-5}$ is used in the denominator to prevent the division-by-zero error. The Parameter-Shared Branched Network (PSBN) in section 2.1 has been trained with separate log dice loss for the disc as well as for the cup. For the weak region of interest model-based segmentation (WRoIM) in section 2.2, similar log dice loss is calculated for training the RoI proposal with respect to the disc mask ground truth and for the final multi-channel output, the average of separate dice losses for the background, cup, and rim is used.

The training data is randomly split with 80% for training and 20% for validation. We apply augmentations by randomly zooming in to a range of $[0.8, 1.2]$, rotating in a range of $[0, 50]$, flipping horizontally and vertically as well as translating in a fractional range of $[0, 0.1]$. The models are trained using SGD optimizer with a learning rate of $10^{-3}$, with momentum in the range of $[0.95, 0.99]$ and a batch size of 1. The best models are saved on the basis of lowest validation loss while training, and the performance of the models are reported on the completely independent test dataset.

## 4. RESULTS AND DISCUSSION

We evaluate and compare the performance of the proposed networks using Dice (Equation 5) and Intersection over Union

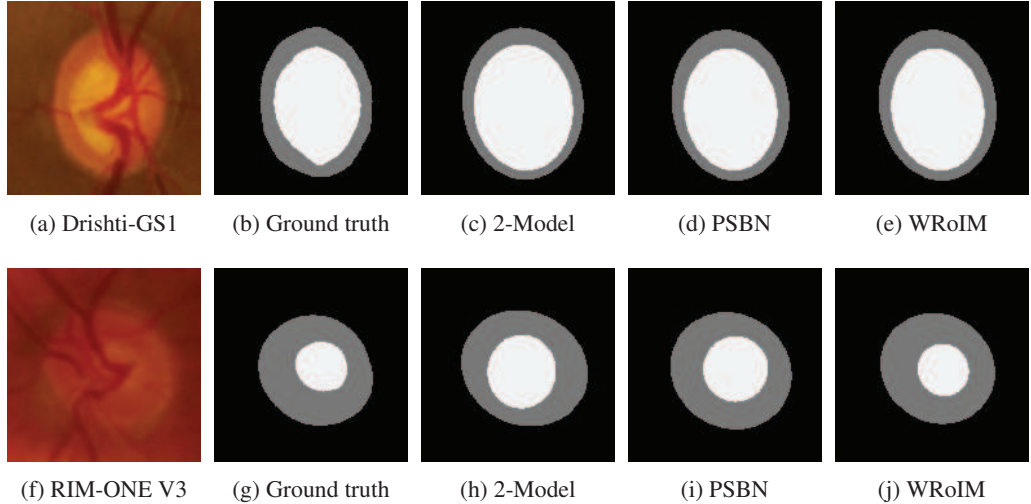

(a) Drishti-GS1  (b) Ground truth  (c) 2-Model  (d) PSBN  (e) WRoIM

(f) RIM-ONE V3  (g) Ground truth  (h) 2-Model  (i) PSBN  (j) WRoIM

**Fig. 4**: Image segmentation comparison on Drishti-GS1 and RIM-ONE v3 sample test images. For clear display of the masks, we have cropped the images. Dice scores for disc/cup (c)-(e) w.r.t. (b) are 0.96/0.88, 0.97/0.91, 0.98/0.95 and for (h)-(j) w.r.t. (g) are 0.93/0.68, 0.94/0.76, 0.97/0.91

| Methods | Parameters | Drishti GS1 | | | | RIM One v3 | | | | Inference time (images/sec) |
|---|---|---|---|---|---|---|---|---|---|---|
| | | Disc | | Cup | | Disc | | Cup | | |
| | | Dice | IOU | Dice | IOU | Dice | IOU | Dice | IOU | |
| Sevastopolsky *et al.* [11] | $19.8 \times 10^6$ | 0.97 | 0.95 | 0.89 | 0.80 | 0.95 | 0.91 | 0.84 | 0.73 | - |
| Sevastopolsky [6] | $1.2 \times 10^6$ | 0.95 | 0.90 | 0.85 | 0.75 | 0.95 | 0.89 | 0.82 | 0.69 | - |
| 2 Model | $15.6 \times 10^6$ | 0.96 | 0.92 | 0.84 | 0.74 | 0.91 | 0.85 | 0.63 | 0.49 | 1.76 |
| Proposed PSBN | $10.9 \times 10^6$ | 0.95 | 0.91 | 0.88 | 0.80 | 0.91 | 0.84 | 0.75 | 0.60 | 0.89 |
| Proposed WRoIM | $7.8 \times 10^6$ | 0.96 | 0.93 | 0.89 | 0.80 | 0.94 | 0.90 | 0.82 | 0.71 | 0.42 |

**Table 2**: Performance comparison of Disc and Cup Segmentation for the proposed and the existing methods

(Equation 6) metrics. A comparative study has also been done with the existing state-of-the-art methods [11, 6] as shown in Table 2. The inference time per image on CPU for different methods is specified as well.

$$IoU(X, Y) = \frac{(X \cap Y)}{(X \cup Y)} \quad (6)$$

For Drishti-GS1, our proposed approach achieves the cup segmentation accuracy close to the previous work [11], while having a lesser number of parameters. For RIM-ONE v3, we are unable to reproduce the results as reported by Sevastopolsky [6]. This could be due to the nature of the dataset, as there is no predefined fixed split between train and test images. Sevastopolsky [6] reports the numbers based on a 5-fold cross-validation, without using any explicit test dataset, while we evaluate on a test set of images that were kept completely aside and not used at all in training. The trend in performance gain of both disc and cup segmentation is consistent with what we obtain for Drishti-GS1 dataset. Our manual-cropping-free approach WRoIM outperforms that of Sevastopolsky [6], in which manual cropping for cup segmentation is done.

## 5. CONCLUSIONS

In this paper, we propose **Parameter-Shared Branched Network** (PSBN) which jointly segments the optic disc and cup from a given fundus image without any prior region of interest (RoI) extraction. The proposed U-Net-based methods uses convolutional layer activations to perform dynamic cropping of the RoI. With the realization that cropping improves the segmentation results, we further developed another architecture **Weak Region of Interest Model-based segmentation** (WRoIM), which first learns the RoI from the fundus image and then jointly segments the optic disc and cup. We show that our method WRoIM uses less number of network parameters while achieving segmentation peformance similar to that of state-of-the-art methods. With lesser parameters, the inference time is shorter for our proposed models than the existing methods. From the results, it is evident that the proposed method(s) can be used as a vital tool(s) in fast and accurate screening of Glaucoma.